\documentclass[10pt,twocolumn,letterpaper]{article}

\usepackage{cvpr}
\usepackage{times}
\usepackage{epsfig}
\usepackage{graphicx}
\usepackage{amsmath}
\usepackage{amssymb}
\usepackage{verbatim} 
\usepackage{bm}
\usepackage{multirow}
\usepackage{colortbl}

 \cvprfinalcopy 

\def\cvprPaperID{****} 

\begin{document}

\title{Factorized Distillation: Training Holistic Person Re-identification Model by Distilling an Ensemble of Partial ReID Models}

\author{Pengyuan Ren\\
SensingTech Ltd.\\
Beijing, China\\
{\tt\small renpengyuan@sensingtech.com.cn}
\and
Jianmin Li\\
Tsinghua University\\
Beijing, China\\
{\tt\small lijianmin@mail.tsinghua.edu.cn}
}

\maketitle

\begin{abstract}
   Person re-identification (ReID) is aimed at identifying the same person across videos captured from different cameras. In the view that networks extracting global features using ordinary network architectures are difficult to extract local features due to their weak attention mechanisms, researchers have proposed a lot of elaborately designed ReID networks, while greatly improving the accuracy, the model size and the feature extraction latency are also soaring. We argue that a relatively compact ordinary network extracting globally pooled features has the capability to extract discriminative local features and can achieve state-of-the-art precision if only the model’s parameters are properly learnt.
In order to reduce the difficulty in learning hard identity labels, we propose a novel knowledge distillation method: Factorized Distillation, which factorizes both feature maps and retrieval features of holistic ReID network to mimic representations of multiple partial ReID models, thus transferring the knowledge from partial ReID models to the holistic network. Experiments show that the performance of model trained with the proposed method can outperform state-of-the-art with relatively few network parameters.

\end{abstract}

\begin{figure}[t]
\begin{center}
\includegraphics[scale=0.3]{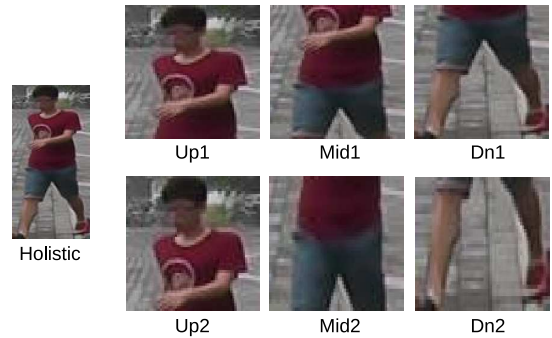}
\end{center}
   \caption{Specification of 7 Views. {\bf Holistic}: pedestrian image resize to 256$\times$128. {\bf Partial Group 1}: Uniformly divide image into 4 stripes to compose Partial Views: Up1(1/4-2/4), Mid1(2/4-3/4),Dn1(3/4-4/4). {\bf Partial Group 2}: Uniformly divide image into 7 stripes to compose Partial Views: Up2(1/7-3/7), Mid2(3/7-5/7), Dn2(5/7-7/7). All partial images resize to 224$\times$224.}
\label{fig:Views}
\end{figure}

\section{Introduction}

Person Re-identification is aimed at identifying the same person across videos captured
from different cameras. It is a challenging task mainly due to factors such as background clutter, pose, illumination and camera point of view variations.  
As the prosperous of deep learning, handcrafted features are replaced by features learned from data by deep convolutional neural networks (CNNs). With feature learning, deep network can build up attention mechanism to reduce interference of background clutter or occlusion, and extract discriminative pose-invariant features. 

It is generally believed that traditional networks extracting globally pooled features (like IDE \cite{IDE}) can only learn to extract salient features during training, and it is hard to extract local features due to their weak attention mechanisms. In order to strengthen attention mechanisms in ReID networks, a lot of approaches have recently been proposed. These works commonly employ spatial partition, body parts detection, pose estimation and so on, pushing the performance of ReID to a new level. However, with the complex network structures or huge models these approaches use, ReID models are hard to be commercially used in a large scale or deployed in mobile devices. Therefore, we want to address the problem in a different way.

Why traditional ReID networks extracting globally pooled features are in low performance? As the number of identities in the training set is small relative to the number of combinations of latent discriminative features, the commonly used identity classification loss prone to overfeat to a small subset of discriminative features which are salient. But identities in testing set are totally different from training set, as a result, the limited features that perfectly classify training identities are not enough to distinguish testing individuals. Even if some visual appearances of body parts in testing images are similiar to local areas in training images, the model still prone to ignore them if they are inconspicuous, because these hard features need not to be leant to lower the loss fucntion.

If we train partial ReID models with high-resolution partial images, each partial model can discover more discriminative features in its restricted region than those found by holistic model in the same area. Each partial ReID model is an expert for a specific region, so the total amount of knowledge contained in the ensemble of separately trained partial ReID models will exceed the holistic model. We aim to transfer the knowledge from multiple partial ReID models to a holistic model, while avoiding feature concatenation in order not to make the dimension of student's retrieval features become unacceptably large when the number of teachers is keep increasing.

Our approach is: 
Firstly, we utilize several separately trained teacher models (holistic or partial) to generate enhanced representation features, named {\bf S}upervisory {\bf R}epresentations ({\bf SR}s).
Secondly, {\bf SR} is regarded as fine-grained high-dimensional soft attributes, and the task of attributes training corresponding to each teacher is added to the holistic student's training system to improve the feature maps of the student.
Thirdly, {\bf SR} is regarded as anchors for each sample in different feature space of partial ReID representation, and the role of metric learning is achieved by factorize  the student model's representation to each of the {\bf SR}'s feature space and mimic these anchors.

The contributions of this paper are:

1) We propose a novel knowledge transfer method, named {\bf F}actorized {\bf D}istillation (FD), which can train a holistic student model by distilling an ensemble of partial models, using the way of factorization instead of feature concatination to result in compact retrieval features even when the number of teachers is large.

2) Trained by FD, ordinary holistic networks extracting global features can also generate strong attention mechanism, and can directly extract partial features from images of high pose variation without incorporating additional body-parts detection or pose estimation networks.

3) Extensive experiments on ReID datasets demonstrate that the proposed method can outperform state-of-the-art with relatively few network parameters.

\section{Related Work}
Person ReID consists of two major techniques: feature engineering/learning and metric learning.
Metric learning makes feature vector to be close within the class and away from other classes\cite{ContrastLoss1,Triplet1}. Feature learning which obsolete feature engineering is to employ deep neural networks to build up attention mechanisms so that feature extraction is not affected by background clutter, and can extract more discriminative features. Previous works have adopted various methods to implement attention mechanisms. These methods can be divided into following categories:

1) Extract local embedding features from pre-defined regions on feature maps. (such as PCB\cite{PCB} and MGN\cite{MGN})

2) Incorporate additional network to generate attention. The generated attention is either as one of inputs to the feature extraction network\cite{PSE,MGCAM,AtnDriven}, or to fuse with feature maps generated by other stream of network\cite{SPReID,PABR,AACN}.,

3) Employ a single network both support attention mechanism and feature extraction, trained jointly with ID labels and regional knowledge.(such as PCB+RPP\cite{PCB})

4) Employ a single network both support attention mechanism and feature extraction, trained only with ID labels, based on specially designed network structures to facilitate the formation of attention mechanisms.\cite{MLFN,HA-CNN,Mancs,AtnDriven}

5) Add attribute labels or text descriptions as supervision to train jointly with ID labels.\cite{IDVR,ARP}

Among above categories, the first category of works (\cite{PCB,MGN}), outperform others on not-precisely-aligned images, by employing simple but effective pre-defined regions, avoiding the instability of error prone attention networks in face of hard samples. We argue that embedding layers after  locally pooled features from pre-defined regions are more prone to overfeat to dataset, because training and testing images share same camera set in main stream ReID datasets. But in practical surveillance scenario where camera angle and positional distribution of detected bounding box relative to body may differ greatly from training dataset, so models extracting embedding features from pre-defined feature map regions may be less robust.

Different from above approaches, our method trains ReID model extracting globally pooled features by means of knowledge distillation.

Knowledge distillation (or knowledge transfer) is a kind of methods employed to transfer pre-digested knowledge (softened supervision) from a teacher network or an ensemble of teachers to student network. the path way of knowledge transfer can be through the output side of the networks\cite{HintonKD} or through the internal hidden layers \cite{FitNets}. At present, knowledge distillation has been widely used in model compression \cite{KD5,KD2,KD7,KD4,KD9,KD6,KD8,KD3,KD1}.

A few of ReID approaches have employed knowledge distillation, such as \cite{DarkRank}, \cite{DML} and \cite{AlignedReID}. \cite{DarkRank} transfer the knowledge of ranking  through deep metric learning. \cite{DML} make an ensemble of students learn collaboratively and teach each other throughout the training process. In addition to classification loss of each student model, \cite{DML} adopt the Kullback Leibler (KL) Divergence between the classification predictions of students. \cite{AlignedReID} developed \cite{DML} by adding batched features distance loss to do metric learning. Different from these methods, we employ an ensemble of partial ReID models as teachers, and transfer knowledge through representations of these teachers with regression losses.

Face verification is a technology field similar to ReID. One of \cite{FMC}'s experiment employed concatenated features as targets which are extracted by an ensemble of regional face models to train the student with regression loss. the concatenated features can be shortened thanks to neuron selection taking use of face attributes dataset. Although neuron selection is very effective, the feature dimension of student is doomed to grow after adding a new teacher to the ensemble, limiting the scalability of the training system. Different from \cite{FMC}, we utilize feature enhancement to reduce intro-identity variance, and use feature factorization to make student's feature dimension do not grow with the number of teachers.

\section{Method}
\begin{figure*}
\begin{center}
\includegraphics[scale=0.35]{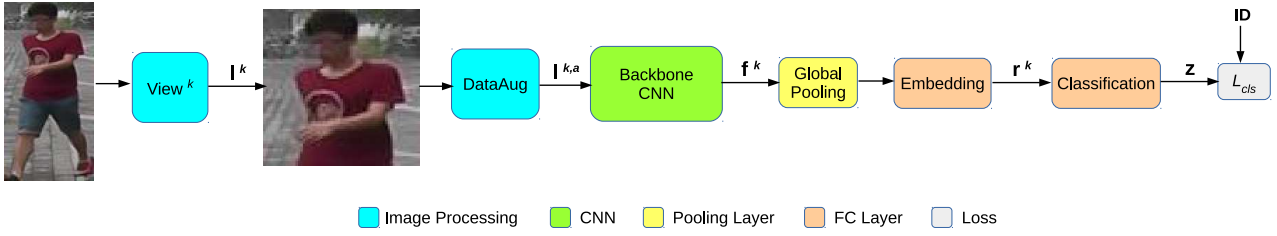}
\end{center}
   \caption{Model training system for a single View. The subscript $i$ of each sample is omitted.}
\label{fig:baseline}
\end{figure*}

\begin{figure*}
\begin{center}
\includegraphics[scale=0.35]{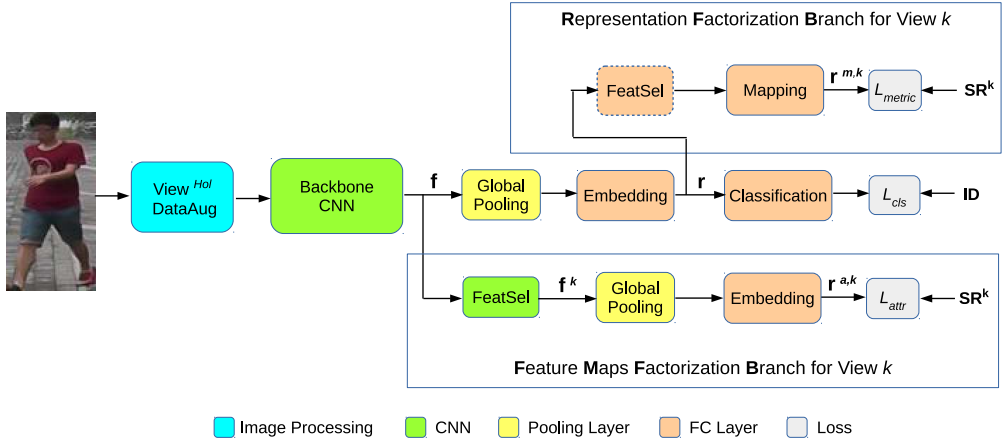}
\end{center}
   \caption{Feature Maps Factorization Branch and Representation Factorization Branch. The subscript $i$ of each sample is omitted}
\label{fig:FMFB+RFB}
\end{figure*}

\subsection{Training ReID Models for Each View}
In this paper, the term ``View'' means a specific area of image as well as the specific resolution for this area to resize. Views can be specified by many ways, such as crop-and-resized images from fixed regions, images of regionally annotated parts, or images of pixel level mask-outed parts. In order to verify our method, we choose the first way as it's the simplest. As such, we define 7 Views shown in Figure \ref{fig:Views}. 

Training multiple partial ReID models for different Views has the following advantages:

1) In order to lower the loss function, partial models must explore more discriminative image features which are not salient in the Holistic View.

2) Crop-and-resized partial image has higher resolution than Holistic View, which can help partial models to discover more subtle image features.

3) Firmly restrict the receptive field of each point in the feature maps of convolutional network. Teaching the holistic student network to extract features from firmly restricted patial areas can make the student adapt to occlusion.

The model training system for a single View is illustrated in Figure \ref{fig:baseline}. We denote a batch of training data as $B=\{{\bf I}_i,y_i\}, i \in [1,N], y_i \in [1,C]$, where ${\bf I}_i$ is the i-th image and $y_i$ is its person ID label, $N$ is batch size. With image processing function ${\rm View}^k$, ${\bf I}_i$ is cropped and resized as the specification for View $k$ to generate ${\bf I}_i^k$, and then generate ${\bf I}_i^{k,a}$ through data augmentation function $\rm DataAug$. ${\bf f}_i^k$ is the feature maps extracted by backbone CNN, and it is processed through a global pooling layer and Embedding block to generate the output representation for ReID. The Embedding block contains a FC (fully connected) layer and a BN (batch normalization) layer. To train the model, a FC layer is added behind ${\bf r}_i^k$ to compute classification logit ${\bf z}_i$, and cross-entropy loss is used, denoted as $L_{cls}$
\begin{equation}
L_{cls} = - \sum_{i=1}^N {\log \frac{ {\bf e}^{{\bf z}_{i,y_i}}} {\sum_{c=1}^C {{\bf e}^{{\bf z}_{i,c}}}}}
\end{equation}
where ${\bf z}_{i,c}$ is sample $i$'s classification logit for ID $c$.

\subsection{Supervisory Representation for Each View}
Knowledge distillation methods generally need to infer teacher models online, but when the number of teachers grows, the computational cost will become prohibitive. In this paper, we generate Supervisory Representations of teachers offline.
Denote teacher model for View $k$ as ${\bm \theta^k}$ and image horizontal flipping function as ${\rm flip(\cdot)}$, Supervisory Representation can be generated by
\begin{equation}
{\bf SR}_i^k = ({\bm \theta}^k ({\rm View}^k({\bf I}_i)) + {\bm \theta}^k({\rm flip}({\rm View}^k({\bf I}_i))/2
\label{equ:SR}
\end{equation}
In above formula, {\bf SR} is generated by averaging features extracted from flipped image and non-flipped image respectively. This is a feature enhancement operation usually used in test time, which can reduce intra class variance of features for each ID. As there is no random operation, {\bf SR} for each sample in the training set can be saved beforehand, and read them while distilling.

Fixed {\bf SR} is a reasonable regularization to  model training. As common image interferences (such as illumination changes and occlusion) can be simulated in data augmentation, if student model is trained with augmented images, and make it mimic fixed {\bf SR} for each sample, thus adding transcendental knowledge to the student, student's adaptability to interference will be significantly improved.

{\bf SR} contains ID-related information (e.g. clothing style, color, sex), while also contains none-ID-related information (e.g. light, occlusion, posture). As the teacher model is trained with heavily augmented data, and generate {\bf SR} through feature enhancement, the none-ID-related information in {\bf SR} extracted from teacher's training set can be significantly eliminated. Therefore, an ensemble of {\bf SR}s has a better feature distribution than that of the student model.

\subsection{Feature Learning with FMFBs}
Our experience is that when training attribute recognition model with globally pooled features, the specific channels in the feature maps preceding global pooling layer will be sensitive only to some local areas, such as head or backpack, even though the attribute labels do not contain positional information. Taking advantage of this phenomenon, \cite{ARP} verified training globally pooled ReID model jointly with attribute recognition losses can improve ReID precision, and result in more semantically interpretable feature maps.

To optimize the feature maps of holistic student model, we take {\bf SR} as human attribute labels. Compared with conventional attribute tags that have tens of dimension and binary values, {\bf SR} is a kind of fine-grained high-dimensional soft labels, carrying much more information.

We use {\bf F}eature {\bf M}aps {\bf F}actorization {\bf B}ranch (FMFB) to imitate an attribute recognition branch for a View (see Figure \ref{fig:FMFB+RFB}). FeatSel block contains a 1$\times$1 convolutional layer, a BN layer and ReLU. It is designed to filter View $k$ relevant information from feature maps ${\bf f}_i$, and output feature maps ${\bf f}_i^k$. 
Embedding block contains a FC layer and a BN layer, taking globally pooled ${\bf f}_i^k$ as input and representation ${\bf r}_i^{{\bf a},k}$ as output (superscript ${\bf a}$ means attribute). We make ${\bf r}_i^{{\bf a},k}$ mimic ${\bf SR}_i^k$
\begin{equation}
L_{attr}^k=\frac {1}{2N} \sum_{i=1}^N{\Vert {\bf SR}_i^k - {\bf r}_i^{{\bf a},k} \Vert_2^2}
\end{equation}

\begin{figure}[t]
\begin{center}
\includegraphics[scale=0.3]{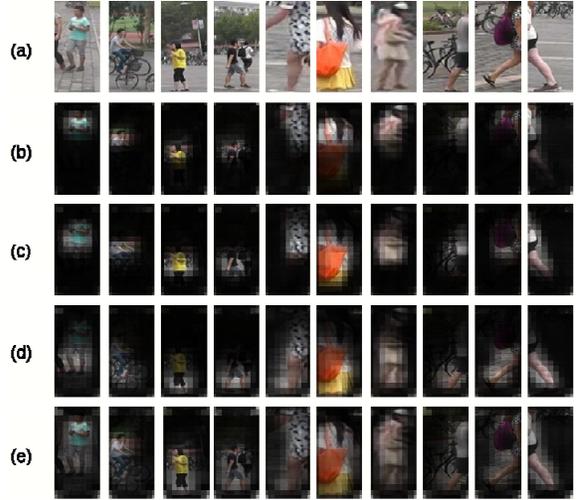}
\end{center}
   \caption{Attentions generated by FMFB. Using L1 norm along channels to get a value on the mask corresponding to each pixel of feature maps, and normalize the the mask to [0,1], then resize it to the same size of original image, and multiply with the original image.
   (a): original image, (b)-(d): ${\bf f}^k$ corresponding to Partial View Group 1, (e): holistic student's ${\bf f}$ optimized by FD.}
\label{fig:attention}
\end{figure}

After training, feature maps ${\bf f}_i^k$ will sensitive to inner area of View $k$. Since there is only a convolutional layer between ${\bf f}_i$ and ${\bf f}_i^k$, the activations of specific channel in ${\bf f}_i$ will be in high correlation with some specific body parts, thus attention mechanism is built (see Figure \ref{fig:attention}).

\subsection{Metric Learning with RFBs}
The Embedding block in holistic student network contains a FC layer, which is by far only optimized by $L_{cls}$. This Embedding block needs metric learning to regularize it, otherwise, it will still be overfited by classification loss. Though conventional metric learning loss functions are contrastive loss\cite{ContrastLoss1} or triplet loss \cite{Triplet1}, in this paper, we propose a new metric learning method using {\bf SR}s as anchors. 

We use {\bf R}epresentation {\bf F}actorization {\bf B}ranch (RFB) (see Figure \ref{fig:FMFB+RFB} ) to factorize the representation. 
The FeatSel block in RBF contains a FC layer, a BN layer and ReLU. Its role is to filter View $k$ relevant information. Note, RFB for Holistic View has no FeatSel block, since both the student and the teacher are in the same View. The Mapping block contains a FC layer and a BN layer. We make ${\bf r}_i^{{\bf m},k}$ mimic ${\bf SR}_i^k$
\begin{equation}
L_{metric}^k=\frac {1}{2N} \sum_{i=1}^N{\Vert {\bf SR}_i^k - {\bf r}_i^{{\bf m},k} \Vert_2^2}
\end{equation}

Our method does not add a loss function directly after the target features as other metric learning methods do. But because there are only one or two FC layers in a RFB, the representation ${\bf r}_i$ of student will also reflect the variance and invariance in each {\bf SR}'s feature space.

\begin{figure}[t]
\begin{center}
\includegraphics[scale=0.32]{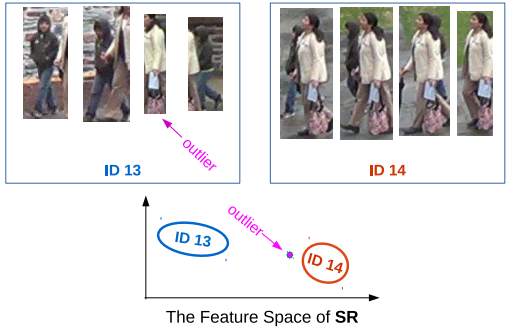}
\end{center}
   \caption{Schematic diagram of outlier.}
\label{fig:metric}
\end{figure}

One advantage of RFB over triplet loss is the adaptability on outlier samples. Figure \ref{fig:metric} shows a typical annotation error in ReID dataset. Outlier sample cause triplet loss generate high value and will make the model overfit to the outlier, which is harmful to model's robustness. But for RFB, due to the outlier's coordinate in {\bf SR} space is close to real ID's sample, the annotation error is automatically eliminated to a large extent.

\subsection{Stabilized Global Max Pooling}
So far, which global pooling adopted in our method has not been introduced. Conventional global pooling methods are  Global Average Pooling (GAP) and Global Max Pooling (GMP). But neither of them fit to our distillation framework. 

If GAP is adopted, in FMFB for a partial View $k$, the activations in feature maps ${\bf f}_k$ outside the region of View $k$ should be suppressed to zeros as much as possible to avoid the predicted representation for View $k$ is interfered by signals from other Views. This regularization is too strict to small student networks.

GMP has the inherent advantage of extracting local features, as a output value of GMP for a specific channel is the maximum value in the feature map. However, GMP has serious shortcomings. For a specific channel, the output of GMP is only relevant to a single point in the feature map, this make it not robust to samples in testing set. In addition, during training, the back-propagational signal of GMP for a channel is also through a single point corresponding to the forward-time maximum. This make parameters of the student model prone to be trapped in the local optimal solution, leading to unstable performance.

In order not only to retain the advantages of GMP, but also to make training stable, and produce high signal-to-noise ratio output during testing, we propose a trick to stabilize GMP:
\begin{equation}
{\bf p}_i={\rm GMP}({\rm AvgPool}({\bf f}_i,kernel={\bf m},stride={\bf 1}))
\label{equ:StableGMP}
\end{equation}
where $\rm GMP(\cdot)$ is conventional Global Max Pooling function, $\rm AvgPool(\cdot)$ is conventional Average Pooling function with kernel size and stride as preset parameters.

\subsection{Put It Together}
The overall training procedure has three steps:

{\bf Step 1}: Train teacher models corresponding to each View and holistic student model, using the training system shown in Figure \ref{fig:baseline}.

{\bf Step 2}: Use all of the teacher models to extract {\bf SR}s from training set, with equation \ref{equ:SR}.

{\bf Step 3}: Add FMFBs and RFBs for all the Views to holistic student's training system, and optimize the student model jointly. The total loss function is
\begin{equation}
L_{total} = L_{cls} + \frac {\alpha}{K}  \sum _{k=1}^K{L_{attr}^k} + \frac {\beta} {K} \sum _{k=1}^K{L_{metric}^k}
\label{equ:TotalLoss}
\end{equation}{}
where $K$ denote the total number of Views, both $\alpha$ and $\beta$ are empirical values.

We adopt the model initialization trick in \cite{InitStudent}. Both weights for feature extraction and ID classification are initialized by pre-trained holistic student model.

\section{Experiments}

\subsection{Implementation Details}
{\bf Specifications of Feature Extraction Networks.} We test ResNet-101 and ResNet-152 respectively as backbone CNN of teachers. The embedding dimension for Holistic View is 512, and for Partial Views is 256. There are 7 student networks to be tested as listed in Tabel \ref{tbl:Students}, varying in backbone CNN and embedding dimension. For ResNet\cite{ResNet}, we reset the stride of conv4$\_$1 from 2 to 1, thus the stride of backbone CNN is 16 pix, same as the stride of SqueezeNet\cite{SqueezeNet}. The size of image as input to the network is specified in the caption of Picture \ref{fig:Views}. The resized image is normalized by each color channel's mean and variance values. The parameter of kernel size in formula \ref{equ:StableGMP} is set to 4. $\alpha$ and $\beta$ in formula \ref{equ:TotalLoss} are 4 and 2 respectively.

{\bf Specifications of FMFBs and RFBs.} There are 7 FMFBs and RFBs corresponding to Holistic View, Partial View Group1 (Up1, Mid1 and Dn1) and Partial View Group2 (Up2, Mid2 and Dn2) (see Figure \ref{fig:Views}).
All the numbers of output channels of FeatSel blocks in FMFBs and RFBs are set to 512. The parameter of kernel size in Equation \ref{equ:StableGMP} is also set to 4.

{\bf Model Training Settings.} We use SGD for training all the models, with batch size as 32, initial learning rate as 0.0025, momentum as 0.9 and decay as 0.0005. In training of the 7 teachers and the initial holistic student model, the learning rates are halved every 20 epochs, and stop training at 80 epochs. In the  training of the final student model, the weights of student and its classification vectors are initialized with pre-trained model, the learning rate is halved every 15 epochs, and stops training at 50 epochs. 

{\bf Data Augmentation Settings.} Data augmentation is very important to the final performance. Ours strategy is to use heavier augmentation for training teacher models and initial student model, and use lighter augmentation for training final student model. The random operations used in each training stage are listed in Table \ref{tbl:DataAug}. Random erasing is described in \cite{RandomErase}. When distilling, if the region of a View is erased by more than 40\%, the regression losses of this sample corresponding to FMFB and RFB are ignored.

{\bf Verification Method.} The features of a sample are the averaged features extracted from flipped image and non-flipped image. Verification score for ranking is calculated by cosine distance between two sample's features.

\begin{table}
\begin{center}
\setlength{\tabcolsep}{1mm}{
\begin{tabular}{|c|c|c|}
\hline\hline
\textbf{Short Name} & \textbf{Network} & \textbf{Enbedding dimension}\\
\hline
S     & SqueezeNet & 512\\
R18a  & ResNet-18  & 512\\
R50a  & ResNet-50  & 512\\
R50b  & ResNet-50  & 2048\\
R101a & ResNet-101 & 512\\
R152a & ResNet-152 & 512\\
R152b & ResNet-152 &2048\\
\hline\hline
\end{tabular}}
\end{center}
\caption{Student networks to be evaluated.}
\label{tbl:Students}
\end{table}

\begin{table}
\begin{center}
\setlength{\tabcolsep}{1mm}{
\begin{tabular}{|c|c|c|c|}
\hline\hline
\textbf{DataAug} & \textbf{Partial} & \textbf{Holistic Teacher} & \textbf{Final}\\
\textbf{(Random)}&\textbf{Teachers}&\textbf{or Initial Student} &\textbf{Student}\\
\hline
\textbf{Flip}        & \checkmark   &      \checkmark          &     \checkmark\\
\textbf{Erasing}     & \checkmark   &      \checkmark          &     \checkmark\\
\textbf{Crop}        & \checkmark   &      \checkmark          &  \\
\textbf{Color}       & \checkmark   &              &  \\
\textbf{Rotation}    & \checkmark   &      & \\
\hline\hline
\end{tabular}}
\end{center}
\caption{Data augmentation methods.}
\label{tbl:DataAug}
\end{table}

\subsection{Datasets and Protocols}
We use three mainstream ReID datasets to evaluate the proposed method. They are:

{\bf Market-1501} A total of six cameras are used, including 5 high-resolution cameras, and one low-resolution camera. This dataset contains 32,668 annotated bounding boxes of 1,501 identities. The pedestrians are cropped with bounding-boxes predicted by DPM detector. The whole dataset is divided into training set with 12,936 images of 751 persons and testing set with 3,368 query images and 19,732 gallery images of
750 persons.

{\bf DukeMTMC-ReID} The whole dataset is captured with 8 high-resolution cameras. Hand-drawn pedestrian bounding boxes are available. There are 1,404 identities appearing in more than two cameras and 408 identities (distractor ID) who appear in only one camera. There are 16,522 training images of 702 identities, 2,228 query images of the other 702 identities and 17,661 gallery images.

{\bf CUHK03} The whole dataset is captured with six surveillance cameras. Each
identity is observed by two disjoint camera views. There are 13,164 images of 1360 identities in the dataset. we adopt the protocol used in \cite{CUHK03-NP}. There are 7,368 training images of 767 identities, 1,400 query images and 5,328 gallery images of 700 identities. We use detected image in our experiments.

In order to facilitate comparisons among combinations of teachers and students, we only evaluate CMC Rank-1 and mAP of models trained separately for each dataset.

\subsection{Evaluation on Different Networks}

\definecolor{mygray}{gray}{.92}
\begin{table}
\begin{center}
\setlength{\tabcolsep}{0.5mm}{
\begin{tabular}{|c|c|cc|cc|cc|}
\hline\hline
\multirow {2}{*}{\textbf{Networks}}& \textbf{Param} & \multicolumn{2}{c|} {\textbf{Market-1501}} & \multicolumn{2}{c|} {\textbf{DukeMTMC}} & \multicolumn{2}{c|} {\textbf{CUHK03}}\\
\cline{3-8} &\textbf{(M)}&Rank-1&mAP&Rank-1&mAP&Rank-1&mAP\\
\hline
S & \multirow {2}{*}{1.0}     & 82.45	& 62.40	& 71.18	& 53.4	& 33.71	& 33.60\\
S+FD &                         & 91.48	& 77.92	& 81.55	& 67.91	& 55.14	& 52.43\\
\hline
R18a & \multirow {2}{*}{11.4}  & 87.83	& 70.73	& 76.84	& 62.22	& 44.29	& 43.00\\
R18a+FD &                       &  93.11	& 82.50 & 85.14 & 73.28	& 64.50	& 61.38\\
\hline
R50a & \multirow {2}{*}{24.6}  & 90.86	& 78.68	& 82.41	& 69.27	& 53.36	& 51.70\\
R50a+FD &                       & 93.65	& 84.38	& 86.40	& 75.44	& 67.79	& 64.51\\
\hline
R50b & \multirow {2}{*}{27.7} & 90.94	& 78.20	& 82.76	& 69.37	& 49.57	& 48.76\\
R50b+FD &                      & 93.68	& 85.22	& 86.49	& 76.07	& 67.71	& 65.24\\
\hline
R101a & \multirow {2}{*}{43.5} & 91.95	& 81.27	& 84.02	& 71.24	& 59.57	& 57.52\\
R101a+FD &                      & 93.76	& 85.22	& 87.07	& 76.32	& 68.57	& 66.33\\
\hline
R152a& \multirow {2}{*}{59.2} & 92.49	& 82.47	& 84.96	& 72.44	& 59.00	& 58.14\\
R152a+FD &                     & 93.76	& 85.62	& 87.12	& 76.80	& 70.79	& 68.37\\
\hline
R152b& \multirow {2}{*}{62.3} & 92.55	& 82.32	& 84.02	& 71.52	& 54.50	& 54.23\\
R152b+FD &                     & 94.33	& 86.95	& 87.88	& 77.74	& 70.86	& 68.21\\
\hline\hline
\rowcolor{mygray}
Holistic &43.5              & 91.95	& 81.27	& 84.02	& 71.24	& 59.57	& 57.52\\
\rowcolor{mygray}
Up1      &43.0              & 73.93	& 47.23	& 74.24	& 55.44	& 32.57	& 33.50\\
\rowcolor{mygray}
Mid1      &43.0             & 76.78	& 56.77	& 68.72	& 49.77	& 32.28	& 33.07\\
\rowcolor{mygray}
Dn1      &43.0              & 71.64	& 52.45	& 64.72	& 45.19	& 32.71	& 30.56\\
\rowcolor{mygray}
Up2      &43.0              & 69.92	& 43.79	& 71.81	& 52.90	& 31.07	& 31.94\\
\rowcolor{mygray}
Mid2      &43.0             & 62.08	& 40.32	& 54.67	& 34.21	& 20.5	& 19.52\\
\rowcolor{mygray}
Dn2      &43.0             & 45.37	& 27.42	& 46.59	& 30.46	& 17.57	& 17.78\\
\hline\hline

\end{tabular}}
\end{center}
\caption{Performance with ResNet-101 teachers.}
\label{tbl:ResNet-101_Teachers}
\end{table}

\begin{table}
\begin{center}
\setlength{\tabcolsep}{0.5mm}{
\begin{tabular}{|c|c|cc|cc|cc|}
\hline\hline
\multirow {2}{*}{\textbf{Networks}}& \textbf{Param} & \multicolumn{2}{c|} {\textbf{Market-1501}} & \multicolumn{2}{c|} {\textbf{DukeMTMC}} & \multicolumn{2}{c|} {\textbf{CUHK03}}\\
\cline{3-8} &\textbf{(M)}&Rank-1&mAP&Rank-1&mAP&Rank-1&mAP\\
\hline
S & \multirow {2}{*}{1.0}         & 82.45	& 62.4	& 71.18	& 53.40	& 33.71	& 33.60\\
S+FD &                             & 91.12	& 77.49	& 82.05	& 67.67	& 54.36	& 52.48\\
\hline
R18a & \multirow {2}{*}{11.4}     & 87.83	& 70.73	& 76.84	& 62.22	& 44.29	& 43.00\\
R18a+FD &                          & 93.14	& 82.27	& 85.23	& 73.09	& 63.07	& 61.02\\
\hline
R50a & \multirow {2}{*}{24.6}     & 90.86	& 78.68	& 82.41	& 69.27	& 53.36	& 51.70\\
R50a+FD &                          & 93.44	& 84.80	& 86.71	& 75.25	& 67.00	& 64.82\\
\hline
R50b & \multirow {2}{*}{27.7}     & 90.94	& 78.20	& 82.76	& 69.37	& 49.57	& 48.76\\
R50b+FD &                          & 94.18	& 85.08	& 87.23	& 76.14	& 66.43	& 64.97\\
\hline
R101a & \multirow {2}{*}{43.5}    & 91.95	& 81.27	& 84.02	& 71.24	& 59.57	& 57.52\\
R101a+FD &                         & 94.09	& 85.51	& 87.21	& 76.02	& 69.79	& 67.41\\ 
\hline
R152a& \multirow {2}{*}{59.2}     & 92.49	& 82.47	& 84.96	& 72.44	& 59.00	& 58.14\\
R152a+FD &                         & 94.12	& 85.76	& 86.80	& 76.51	& 70.07	& 67.85\\    
\hline
R152b& \multirow {2}{*}{62.3}     & 92.55	& 82.32	& 84.02	& 71.52	& 54.50	& 54.23\\
R152b+FD &                         & 95.07	& 87.10	& 88.06	& 77.64	& 70.00	& 68.04\\   
\hline\hline
\rowcolor{mygray}
Holistic &59.2                    & 92.49	& 82.47	& 84.96	& 72.44	& 59.00	& 58.14\\
\rowcolor{mygray}
Up1      &58.7                    & 73.19	& 47.77	& 74.06	& 55.19	& 32.86	& 33.66\\
\rowcolor{mygray}
Mid1      &58.7                   & 76.57	& 57.42	& 69.84	& 49.41	& 33.79	& 32.88\\
\rowcolor{mygray}
Dn1      &58.7                    & 74.14	& 54.32	& 65.13	& 45.97	& 32.71	& 30.48\\
\rowcolor{mygray}
Up2      &58.7                    & 69.51	& 43.92	& 74.19	& 54.60	& 32.14	& 33.81\\
\rowcolor{mygray}
Mid2      &58.7                   & 63.39	& 41.87	& 54.58	& 34.77	& 20.29	& 20.23\\
\rowcolor{mygray}
Dn2      &58.7                    & 46.50	& 27.94	& 49.24	& 31.70	& 19.21	& 18.67\\
\hline\hline

\end{tabular}}
\end{center}
\caption{Performance with ResNet-152 teachers.}
\label{tbl:ResNet-152_Teachers}
\end{table}

The performances on ResNet-101 teachers are shown in Table \ref{tbl:ResNet-101_Teachers}, and the performances on ResNet-152 teachers are shown in Table \ref{tbl:ResNet-152_Teachers}. In each table, the number of parameters of student networks increases from less 1 million (SqueezeNet) to more than 60 million (ResNet-152).

It can be found in the two tables that basically performance increases with number of parameters. Although all of the three datasets are very small, big models based on deep residual networks do not overfit to the training IDs.

All of the students see significant improvements after {\bf FD} are utilized. For student network of ResNet-152 (R152b) optimized by ResNet-101 teachers, although the performance of each teacher is much lower than the student, (Rank-1,mAP) of the stronger student are still increased by (1.78,4.63), (3.86, 6.22) and (16.36, 13.98) respectively on the three datasets. This verified that: Firstly, training specialized teacher models for different Views can discover and aggregate more knowledge from dataset, though each teacher is very weak when it is used individually. Secondly, the proposed distilling method can pumping knowledge from an ensemble of specialized but weaker models, and transfer the knowledge to a already strong model and make it even stronger.

It can be seen from the relationship between performance enhancements and model parameter sizes, that {\bf FD} is more effective to smaller students. For ResNet-18, the smallest ResNet model with 11.4 M parameters, after optimized by ResNet-101 teachers , (Rank-1,mAP) are increased by (5.28, 11.77), (8.3, 11.06) and (20.21, 18.38), which are more significant than that of bigger student networks achieved. While, for SqueezeNet, whose parameter quantity is 11$\times$ less than ResNet-18, achives (Rank-1, mAP) improvements of (9.03, 15.52), (10.37, 14.51) and (21.43, 18.83), which are the highest increasements among all the students. The big improvement on SqueezeNet student also indicate that the effectiveness of Factorized Distillation is not limited to Residual Networks. The performance of SqueezeNet+FD is comparable to baseline ResNet-50 model, so it can be interpreted as compressing a baseline ReID model by 25$\times$ without obvious precision losses. 

In comparison of the performance improvements achieved by different embedding dimensions (512-Dim vs 2048-Dim on ResNet-50 and on ResNet-152), we found higher embedding dimension can result in more performance improvements in the majority of cases. But there are also exceptions. For CUHK03, performances of students with 2048-Dim are generally less than that of 512-Dim. We speculate there are two reasons: Firstly, higher embedding dimension lead to the number of parameters of FC layer for ID classification increases accordingly, which needs more training samples to get rid of overfit, but the training set of CUHK03 is much smaller than Market-1501 and DukeMTMC-ReID. Secondly, one characteristic of CUHK03 is that, same individual is only appears in two cameras with large direction difference. This characteristic may hinder discovering more discriminative features.

In comparison of the two tables, we found that bigger teachers (ResNet-152) can get higher performance improvements on bigger students or larger training datasets, as for the improvements on student networks bigger than ResNet-18 and datasets other than CUHK03, improvements made by ResNet-152 teachers are higher than that made by ResNet-101 teachers. While, ResNet-101 teachers are good at smaller students or smaller dataset. We speculate that there are two reasons for this phenomenon: Firstly, the {\bf SR} generated by ResNet-152 may be too strict for small students to mimic and needs more samples to train, which is not satisfied by CUHK03. Secondly, off-line saved {\bf SR} may not include all the information of the teacher model, even though fixed {\bf SR} can introduce transcendental knowledge to the student. From this perspective, teacher models should not be made as big as possible.  


\subsection{Comparison with State-of-the-Art}

\begin{table}
\begin{center}
\setlength{\tabcolsep}{0.1mm}{
\begin{tabular}{|c|c|cc|cc|cc|}
\hline\hline
\multirow {2}{*}{\textbf{Model}}& \textbf{Param} & \multicolumn{2}{c|} {\textbf{Market-1501}} & \multicolumn{2}{c|} {\textbf{DukeMTMC}} & \multicolumn{2}{c|} {\textbf{CUHK03}}\\
\cline{3-8} &\textbf{(M)}&Rank-1&mAP&Rank-1&mAP&Rank-1&mAP\\
\hline
\multicolumn{8}{|c|} {Param. $<$ 10 M}\\
\hline
AACN\cite{AACN} & 8.0	&85.9	&66.87	&76.84	&59.25 & $-$ & $-$\\
NIN-BN+ & \multirow {2}{*}{7.6} & \multirow {2}{*}{86.7} &\multirow {2}{*}{68.2} & \multirow {2}{*}{$-$} &\multirow {2}{*}{$-$} &\multirow {2}{*}{$-$} &\multirow {2}{*}{$-$}\\
DarkRank\cite{DarkRank} &&&&&&&\\
MobileNet & \multirow {2}{*}{3.3} & \multirow {2}{*}{87.73} &\multirow {2}{*}{68.83} & \multirow {2}{*}{$-$} &\multirow {2}{*}{$-$} &\multirow {2}{*}{$-$} &\multirow {2}{*}{$-$}\\
+DML\cite{DML} &&&&&&&\\
HA-CNN\cite{HA-CNN} & 2.7 & 91.2	& 75.7	& 80.5	& 63.8	& 41.7	& 38.6\\
\hline
S+FD   &1.0  &\textbf{91.48}	&\textbf{77.92}	&\textbf{81.55}	&\textbf{67.91}	&\textbf{55.14}	&\textbf{52.43}\\
\hline
\multicolumn{8}{|c|} {20 M $<$ Param. $<$ 30 M}\\
\hline
PSE\cite{PSE}   & 27.0  & 87.7	& 69.0	    &79.8	&62.0     &{$-$}  &{$-$}\\
DaRe\cite{RA}    & 28.7	& 89.0	& 76.0	& 80.2	& 64.5	& 63.3	& 59.0\\
SPReID\cite{SPReID}  & 27.1	& 92.54	& 81.34	& 84.43	& 70.97 &{$-$}  &{$-$}\\
SL\cite{SL}    & 29.8	& 89.43	& 72.58	& 73.58	& 53.2  &{$-$}  &{$-$}\\
PCB+RPP\cite{PCB}   & 26.6	& 93.8	& 81.6	& 83.3	& 69.2	& 63.7	& 57.5\\
Mancs\cite{Mancs} & 27.9	& 93.1	& 82.3	& 84.9	& 71.8	& 65.5	& 60.5\\
\hline
R50a+FD & 24.6	& 93.44	& 84.8	& 86.71	& 75.25	&\textbf{67.79}	& 64.51\\
R50b+FD & 27.7     & \textbf{94.18} & \textbf{85.08} & \textbf{87.23}	& \textbf{76.14} & 67.71 & \textbf{65.24}\\
\hline
\multicolumn{8}{|c|} {Param. $>$ 60 M}\\
\hline
MGN\cite{MGN}   &70.3	&\textbf{95.7}	&86.9	&\textbf{88.7}	&\textbf{78.4}	&66.8	&66\\
R152b+FD &62.3	&95.07	&\textbf{87.1}	&88.06	&77.64	&\textbf{70.86}	&\textbf{68.21}\\

\hline\hline 

\end{tabular}}
\end{center}
\caption{Comparison with state-of-the-art. S+FD models corresponding to each dataset and the models for CUHK03 are distilled by ResNet-101 teachers, other models are distilled by ResNet-152 teachers.}
\label{tbl:Comparison}
\end{table}

\begin{table}
\begin{center}
\setlength{\tabcolsep}{0.5mm}{
\begin{tabular}{|c|cc|cc|}
\hline\hline
\multirow {2}{*}{\textbf{Model}} & \multicolumn{2}{c|} {\textbf{Market-1501}} & \multicolumn{2}{c|} {\textbf{DukeMTMC}}\\
\cline{2-5}&Rank-1&mAP&Rank-1&mAP\\
\hline
PUL\cite{PUL}(\textbf{CUHK03}) &41.9	&18	&23	&12 \\
PUL\cite{PUL}(\textbf{Market}) &{$-$}  &{$-$} &30.4 &16.4\\
SPGAN\cite{SPGAN} &	51.5	&22.8	&41.1	&22.3\\
\hline
R50b(\textbf{CUHK03})	    & 47.65	    &24.38	&32.81	&18.37\\
R50b+FD(\textbf{CUHK03})	&\textbf{54.31}	&\textbf{30.88}	&38.73	&22.96\\
R50b+FD(\textbf{Market})	&{$-$}  &{$-$}  &\textbf{44.75}	&\textbf{26.97}\\
\hline\hline 

\end{tabular}}
\end{center}
\caption{Cross dataset comparison. Different form listed others, we don't use target dataset to finetune our model.}
\label{tbl:CrossDataset}
\end{table}

In this section, we compare our performances on the three datasets with recently proposed state-of-the-art approaches. The comparison is divided into three groups according to number of parameters. It makes the comparison fair since number of parameters significantly affect performance. The parameter quantity of other methods are either get from their papers or inferred from their implementation details. Although the number of parameters inferred may has slight error, the grouping of the method is not affected.   

As shown in Table \ref{tbl:Comparison}, our method outperform state-of-the-art in the group of {\bf Param $<$ 10 M} and {\bf 20 M $<$ Param $<$ 30 M}, especially in the first group, S+FD's parameter size is 2.7$\times$ less than HA-CNN\cite{HA-CNN}. In the Group of {\bf Param $>$ 60 M}, there are only two models: MGN\cite{MGN} and our ResNet-152+FD. ResNet-152+FD outperform MGN on CUHK03 by a large margin even though our model has 8 million less parameters.

Thanks to Stabilized GMP that inherently adapt to different domains to some extent, the cross-dataset direct transfer performances are competitive compaired with domain adaptation methods\cite{SPGAN,PUL} recently proposed. (see Table \ref{tbl:CrossDataset})


\subsection{Ablation Analysis}
\begin{table}
\begin{center}
\setlength{\tabcolsep}{0.2mm}{
\begin{tabular}{|c|cc|cc|cc|}
\hline\hline
R50b & \multicolumn{2}{c|} {\textbf{Market-1501}} & \multicolumn{2}{c|} {\textbf{DukeMTMC}} & \multicolumn{2}{c|} {\textbf{CUHK03}}\\
\cline{2-7}+$LossX$&Rank-1&mAP&Rank-1&mAP&Rank-1&mAP\\
\hline
$L_{cls}$ & 90.94	& 78.2	& 82.76	& 69.37	& 49.57	& 48.76\\
$L_{cls}+L_{attr}$ & 93.32	& 84.69	& 85.77	& 75.47	& 64.43	& 62.37\\
$L_{cls}+L_{metric}$ & 92.96	& 83.76	& 87.07	& 75.39	& 63.79	& 62.09\\
$L_{cls}+L_{attr}$ & \multirow {2}{*}{94.18}	& \multirow {2}{*}{85.08}	& \multirow {2}{*}{87.23}	& \multirow {2}{*}{76.14}	& \multirow {2}{*}{67.71}	& \multirow {2}{*}{65.24}\\
$+L_{metric}$ & & & & & &\\
\hline\hline 
\end{tabular}}
\end{center}
\caption{Analysis of losses.}
\label{tbl:Ablation1}
\end{table}

\begin{table}
\begin{center}
\setlength{\tabcolsep}{0.2mm}{
\begin{tabular}{|c|cc|cc|cc|}
\hline\hline
R50b & \multicolumn{2}{c|} {\textbf{Market-1501}} & \multicolumn{2}{c|} {\textbf{DukeMTMC}} & \multicolumn{2}{c|} {\textbf{CUHK03}}\\
\cline{2-7}+Teacher$X$&Rank-1&mAP&Rank-1&mAP&Rank-1&mAP\\
\hline
None & 90.94	& 78.2	& 82.76	& 69.37	& 49.57	& 48.76\\
Hol & 92.1	& 82.64	& 85.32	& 73.54	& 58.79	& 57.7\\
Hol+PartialG1 & 93.82	& 84.4	& 86.67	& 75.69	& 64.36	& 63.23\\
Hol+PartialG1 & \multirow {2}{*}{94.18}	& \multirow {2}{*}{85.08}	& \multirow {2}{*}{87.23}	& \multirow {2}{*}{76.14}	& \multirow {2}{*}{67.71}	& \multirow {2}{*}{65.24}\\
+PartialG2 & & & & & & \\
\hline\hline 
\end{tabular}}
\end{center}
\caption{Analysis of teachers.}
\label{tbl:Ablation2}
\end{table}

In this section, we analysis the effectiveness of each part of Factorized Distillation. 

It can be found from Table \ref{tbl:Ablation1} that both {\bf FMFB} and {\bf RFB} cause considerably improvements, and can complement each other when jointly used.

In Table \ref{tbl:Ablation2}, we compare effectiveness of different teachers. {\bf Hol} denote the teacher of Holistic View, {\bf PartialG1} denote teachers of Partial Group 1 (Up1,Mid1 and Dn1) and {\bf PartialG2} denote teachers of Partial Group 2 (Up2, Mid2 and Dn2). This experiment shows the more teachers we employ, the more improvement we get.

\section{Conclusion}

In this paper, we proposed Factorized Distillation, a method of training holistic ReID model by distilling multiple patial ReID models. 
Factorized Distillation is friendly to add on more teachers, so it is very scalable. And it is a flexible framework that can be implemented in many ways. Due to the limitation of time and page number, this paper only illustrate a special case of FD. Below are a few other variants that may also be tried, such as:
a) Add teachers that trained with annotated regions or masks of body parts.
b) Replace the simple teacher networks in this paper with other state-of-the-art networks.
c) Generate {\bf SR}s online, though this will cost much more computing resources.


{\small
\bibliographystyle{plain}
\bibliography{mybib}
}

\end{document}